\documentclass[lettersize,journal]{IEEEtran}
\usepackage{amsmath,amsfonts}
\usepackage{algorithmic}
\usepackage{algorithm}
\usepackage{array}
\usepackage[caption=false,font=normalsize,labelfont=sf,textfont=sf]{subfig}
\usepackage{textcomp}
\usepackage{stfloats}
\usepackage{url}
\usepackage{verbatim}
\usepackage{graphicx}
\usepackage{cite}

\usepackage{natbib}
\setcitestyle{numbers,square}
\usepackage{amsmath}
\usepackage{amssymb}
\usepackage{xcolor}
\usepackage{booktabs}
\usepackage{orcidlink}
\usepackage{multirow}
\usepackage{bbding}
\usepackage{orcidlink}
\usepackage{bbm}
\usepackage{natbib}
\usepackage[capitalise]{cleveref}
\usepackage{hyperref}
\usepackage{url}
\usepackage{booktabs}
\usepackage{multirow}
\usepackage{cleveref}
\usepackage{amsmath}
\usepackage{amsthm}
\usepackage{amsmath,amssymb,amsthm}
\usepackage{mathtools}
\usepackage{tcolorbox}
\usepackage{indentfirst}
\usepackage[table]{xcolor}

\begin{document}

\title{The Structural Scalpel: Automated Contiguous Layer Pruning for Large Language Models}
% \title{CLP: Continuous Layer Pruning via Differentiable Mask and Targeted Recovery}

% \author{Anonymous
% % IEEE Publication Technology,~\IEEEmembership{Staff,~IEEE,}
%         % <-this % stops a space
% \thanks{This paper was produced by the IEEE Publication Technology Group. They are in Piscataway, NJ.}% <-this % stops a space
% \thanks{Manuscript received April 19, 2021; revised August 16, 2021.}}

\author{Yao Lu\orcidlink{0000-0003-0655-7814},~\IEEEmembership{Student Member, IEEE}, Yuqi Li, Wenbin Xie, Shanqing Yu, Qi Xuan\orcidlink{0000-0002-6320-7012},~\IEEEmembership{Senior Member,~IEEE}, Zhaowei Zhu, Shiping Wen\orcidlink{0000-0001-8077-7001}
        % <-this % stops a space
\thanks{This work was partially supported by the Key R\&D Program of Zhejiang under Grant 2022C01018 and by the National Natural Science Foundation of China under Grant U21B2001, 62301492 and 61973273.}% <-this % stops a space
% \thanks{Maoyu Wang is with the College of Computer science and Technology, Zhejiang University of Technology, Hangzhou, China (e-mail: wangmy.zjut@gmail.com)}
\thanks{Yao Lu is with the Institute of Cyberspace Security, College of Information Engineering, Zhejiang University of Technology, Hangzhou 310023, China, with the Binjiang Institute of Artificial Intelligence, Zhejiang University of Technology, Hangzhou 310056, China, also with the Centre for Frontier AI Research, Agency for Science, Technology and Research, Singapore 138632 (e-mail: yaolu.zjut@gmail.com).}
\thanks{Yuqi Li is with the City College of New York, City University of New York, USA (e-mail: yuqili010602@gmail.com).}
\thanks{Wenbin Xie, Shanqing Yu and Qi Xuan are with the Institute of Cyberspace Security, College of Information Engineering, Zhejiang University of Technology, Hangzhou 310023, China, also with the Binjiang Institute of Artificial Intelligence, Zhejiang University of Technology, Hangzhou 310056, China (e-mail: xavierxie948@gmail.com, yushanqing@zjut.edu.cn, xuanqi@zjut.edu.cn).}
\thanks{Shiping Wen is with the Australian AI Institute, Faculty of Engineering and Information Technology, University of Technology, Australia (e-mail: Shiping.Wen@uts.edu.au).}
\thanks{Zhaowei Zhu is with the Binjiang Institute of Artificial Intelligence, Zhejiang University of Technology, Hangzhou 310056, China (e-mail: zwzhu1995@gmail.com).}
}

% The paper headers
\markboth{Journal of \LaTeX\ Class Files,~Vol.~14, No.~8, August~2021}%
{Shell \MakeLowercase{\textit{et al.}}: A Sample Article Using IEEEtran.cls for IEEE Journals}

% \IEEEpubid{0000--0000/00\$00.00~\copyright~2021 IEEE}
% Remember, if you use this you must call \IEEEpubidadjcol in the second
% column for its text to clear the IEEEpubid mark.

\maketitle

\begin{abstract}
Although large language models (LLMs) have achieved revolutionary breakthroughs in many fields, their large model size and high computational cost pose significant challenges for practical deployment on resource-constrained edge devices. To this end, layer pruning has been proposed to reduce the computational overhead by directly removing redundant layers. However, existing layer pruning methods typically rely on hand-crafted metrics to evaluate and remove individual layers, while ignoring the dependencies between layers. This can disrupt the model's information flow and severely degrade performance. To address these issues, we propose CLP, a novel continuous layer pruning framework that introduces two key innovations: a differentiable concave gate algorithm that automatically identifies the best continuous layer segments for pruning via gradient-based optimization; and a cutoff endpoint tuning strategy that effectively restores model performance by fine-tuning only the layers adjacent to the pruned segments. Extensive experiments across multiple model architectures (including LLaMA2, LLaMA3 and Qwen) and sizes (from $7$B to $70$B parameters) show that CLP significantly outperforms existing state-of-the-art baselines. For example, at a pruning rate of $20\%$, CLP achieves an average performance retention of $95.34\%$ on LLaMA3-70B, outperforming baselines by $4.29\%$-$30.52\%$. Furthermore, CLP can be seamlessly combined with quantization to further compress the model with only a slight performance loss.
\end{abstract}


\section{Introduction}
\IEEEPARstart{I}{n} recent years, large language models (LLMs), notably the GPT~\citep{floridi2020gpt,achiam2023gpt}, LLaMA~\citep{touvron2023llama,grattafiori2024llama}, and Qwen~\citep{yang2024qwen2,yang2025qwen3} families, have achieved revolutionary breakthroughs in natural language processing. These models, with their powerful emergent capabilities, have achieved performance close to or even surpassing human performance in a wide range of tasks, such as machine translation~\citep{xu2024contrastive,enis2024llm}, mathematical analysis~\citep{zhang2024mathverse,ahn2024large}, code generation~\citep{liu2024exploring,wang2024enhancing}, complex reasoning~\citep{wang2023can,wang2025see,leong2025should,yi2025score,li2025multi}, and open dialogue~\citep{yi2024survey,mendoncca2024benchmarking}. However, their exceptional performance comes at a significant computational cost. The sheer number of parameters not only demands massive storage but also incurs high inference latency and substantial energy consumption, severely hindering deployment of LLMs on resource-constrained platforms such as mobile and edge devices. 

To address this challenge, numerous techniques~\citep{xu2024survey,liu2024spinquant,ma2023llm,gao2025dspc} have been developed to create more efficient and compact models. Among these methods, model pruning, as a technique for directly removing redundant parts of the model, has attracted much attention due to its efficiency and effectiveness. Depending on the granularity of the pruning target, pruning methods can be roughly divided into weight pruning~\cite{frantar2023sparsegpt,sun2023simple}, channel pruning~\cite{ma2023llm,zhang2025leank}, and layer pruning~\cite{lu2024reassessing,men2024shortgpt}. Weight pruning is the most fine-grained method, which removes individual weights to obtain a sparse parameter matrix. However, this often leads to irregular sparsity and requires specialized hardware for effective acceleration~\cite{ling2024slimgpt}. Channel pruning, which performs structured trimming in the channel dimension, maintains hardware friendliness but may disrupt key attention patterns. In contrast, layer pruning directly removes entire network layers, maximizing computational efficiency while maintaining the integrity of the original model's structure. Therefore, we focus on layer pruning methods in this paper.

Existing studies on layer pruning focuses primarily on evaluating and removing unimportant layers. For example, some methods calculate an independent importance score for each layer based on weight norms~\cite{lu2024alphapruning,kim2024shortened} and feature representations~\cite{men2024shortgpt,liu2024foldgpt}, and then remove the layers with the lowest scores. Others~\cite{zhang2024finercut,song2024sleb} utilize search algorithms to identify subnetworks with superior performance. While these methods have achieved promising results, they often evaluate each layer in isolation, ignoring the model's inter-layer dependencies and structural integrity. These strategies often result in the removal of multiple non-contiguous layers, severely disrupting the smooth information flow of features~\cite{zhang2024unified,dutta2021redesigning} within the model. To address the above issues, we propose a novel \textbf{C}ontinuous \textbf{L}ayer \textbf{P}runing (\textbf{CLP}) method, which aims to automatically discover and remove consecutive layer segments that have the least impact on model performance, thereby achieving efficient and stable model compression.

Specifically, the CLP framework consists of two core components that work together to form a complete end-to-end pruning solution. Firstly, we propose a differentiable concave gating algorithm to automatically locate the optimal continuous pruning region, which overcomes the limitation of traditional methods that isolate the importance of individual layers. Subsequently, we use the KL divergence between the original and pruned model outputs as the optimization objective, and employ gradient descent to automatically search for continuous layers that have the least impact on overall performance. Secondly, after removing consecutive layers, we develop a unique cutoff endpoint tuning strategy for efficient performance recovery. Unlike traditional full fine-tuning or LoRA fine-tuning~\cite{hu2022lora}, our strategy focuses optimization on the adjacent layers on both sides of the removed segment (i.e., the "cutoff endpoints"). By fine-tuning the parameters of only these two key layers, we can accurately "stitch" the structural breaks caused by pruning with minimal computational cost, significantly improving the efficiency of performance recovery. Extensive systematic experiments demonstrate the effectiveness of our approach on multiple benchmarks and models of different sizes and architectures.

In a nutshell, we make the following contributions:
\begin{itemize}
    \item We propose a differentiable concave gating algorithm that automates the search for pruning regions by constructing a continuously differentiable soft mask. The algorithm optimizes the KL divergence between the outputs of the original and pruned models and utilizes gradient descent to automatically locate consecutive layers that have the least impact on model performance, avoiding the need to evaluate individual layers in isolation as in traditional pruning methods.
    \item We introduce a novel performance recovery strategy called cutoff endpoint tuning. Unlike the most widely used LoRA fine-tuning, this mechanism only fine-tunes the full parameters of the key layers adjacent to the ends of the pruned segment. This strategy can repair the structural damage caused by pruning at a very low computational cost.
    \item We conduct extensive experiments on models of various sizes and architectures (such as LLaMA2, LLaMA3 and Qwen) and multiple benchmark datasets. The results consistently show that our approach outperforms state-of-the-art baselines in terms of average performance retention. Besides, CLP can be seamlessly integrated with mainstream quantization techniques, achieving further extreme compression with only minimal performance loss. This provides a practical technical path for deploying LLMs on resource-constrained edge devices.

\end{itemize}

\section{Related Work}
\textbf{Layer Pruning.} Unlike fine-grained pruning by weight~\citep{sun2023simple} or channel~\citep{ma2023llm,chen2023rgp,li2025sepprune}, layer pruning directly removes the entire layer to reduce computational complexity. It significantly reduces inference latency and memory usage without changing the tensor dimension, making it easier to deploy. Consequently, much research has focused on introducing effective layer pruning methods. For example, ShortGPT~\citep{men2024shortgpt} uses a BI score to assess layer importance and remove less important layers. Similarly, Gromov et al.~\cite{gromov2024unreasonable} utilize angular distance to to determine and remove less important layers, and employs QLoRA to restore performance. LaCo~\citep{yang2024laco} rears model layers collapse into a prior layer to quickly reduce the model size. SLEB~\citep{song2024sleb} uses the perplexity to calculate the layer importance and iteratively discard those unimportant layers. Recently, Lu et al.~\cite{lu2024reassessing} conduct extensive experiments across $7$ layer pruning metrics and find that simply pruning the last several layers performs better than many complex pruning metrics. Unlike traditional layer pruning methods, CLP automatically discovers the optimal contiguous blocks of layers to remove through gradient-based mask learning. Furthermore, unlike existing pruning methods~\cite{kim2024shortened,men2024shortgpt} that rely on LoRA fine-tuning, CLP employs targeted cutoff endpoint tuning, focusing its efforts on restoring the performance of pruned models on layers adjacent to the pruned segments. This fine-tuning strategy not only improves the accuracy of the pruned model and reduces the computational overhead, but more importantly, it establishes a new paradigm for fine-tuning the pruned model.

\textbf{Supervised Finetuning.} Supervised fine-tuning is one of the main paradigms for adapting pre-trained models to downstream tasks. It improves the performance of the model on the target task by performing supervised gradient~\cite{zhang2025beyond} updates on model parameters on labeled data. Traditional methods typically require end-to-end fine-tuning of all parameters, which is costly. To address this issue, researchers have proposed a variety of parameter-efficient fine-tuning strategies, such as LoRA~\citep{hu2022lora,dettmers2023qlora,zhang2025sensitivity} and Adapter-Tuning~\citep{houlsby2019parameter,wang2022adamix,leong2024efficient}. By updating only a small number of parameters or inserting lightweight modules, these strategies significantly reduce training and storage costs while maintaining performance. Given the superiority of LoRA in practice, many mainstream pruning methods~\citep{ma2023llm,men2024shortgpt,kim2024shortened} choose LoRA as the preferred strategy for performance recovery of pruned models. Different from existing pruning methods, our cutoff endpoint tuning adopts a highly targeted fine-tuning strategy. Specifically, after removing consecutive layers, we perform full parameter fine-tuning specifically on the preceding and succeeding layers of the pruned segment. This design effectively restores the information flow disrupted by the removal of consecutive layers. By precisely focusing updates on these critical layers, our approach achieves more effective and computationally efficient performance recovery than widely used but less targeted fine-tuning methods.

\section{Methodology}
\label{sec:method}

In this section, we introduce CLP, a continuous layer pruning algorithm based on a fully differentiable concave gating mechanism. The core of CLP is to dynamically remove consecutive layers using smoothly parameterized gating, while strictly limiting the difference between the pruned model's output and the original model's output. After obtaining the pruned model, we perform cutoff endpoint tuning to restore the performance.

\subsection{Problem Setup}
Given a pre-trained LLM $\mathcal{H}=f_L\circ f_{L-1}\circ\cdots\circ f_1$, which consists of $L$ consecutive transformer layers. Let the mapping of $f_i$ layer be
\begin{equation}
  o_i= \mathrm{TransformerLayer}_i\bigl(o_{i-1}\bigr), 
  \quad i=1,\dots,L,
  \label{eq:standard-block}
\end{equation}
where $x$ is the input data and $o_i$ denotes the output of the $i$th layer. Our goal is to remove $n$ consecutive layers from the pre-trained model to obtain a more compact model while preserving its performance as much as possible. Mathematically, this problem can be formalized as follows:
\begin{equation}
\begin{aligned}
\min_{a\in\{1,\dots,L-n\}}\; 
&\mathbb{E}_{(x,y)\sim\mathcal{D}}\big[\mathcal{L}\big(\mathcal{H}_{\mathrm{prune},a}(x),\,y\big)\big] \\
\quad\text{s.t.} \, \mathcal{H}_{\mathrm{prune},a}
&:=f_L \circ \cdots\circ f_{a+n}\circ f_{a-1}\circ \cdots\circ f_1,
\end{aligned}
\label{eq:pruning normal loss}
\end{equation}
where \(a\in[1,L-n]\) is the starting index of the pruned layers, \(n\in[1,L]\) is the number of pruned layers, $\mathcal{D}$ denotes the target dataset and $\mathcal{L}(\cdot)$ is a task-specific loss function.

\subsection{Differentiable Concave Gating Algorithm}
To solve this problem, an intuitive approach is to directly brute-force enumerate all possible $(a,n)$ combinations in discrete space. However, such method has two significant drawbacks: (i) the search space grows quadratically with the number of layers, making the enumeration cost prohibitive when the number of layers $L$ is large. (ii) each set of candidate pruning configurations requires a complete model performance test, resulting in high computational cost.
To address these issues, we introduce a layer-wise differentiable mask vector $\mathcal{M} = [m_1, m_2, \dots, m_{L}]$, where 
% To obtain the optimal sequence of consecutive pruned layers, we introduce a layer-wise differentiable mask vector $\mathcal{M} = [m_1, m_2, \dots, m_{L}]$, where 
\begin{equation}
    m_i =
  \begin{cases}
    0, & a \le i \le a + n-1\\
    1, & \text{otherwise}.
  \end{cases}
  \label{eq:mask}
\end{equation}
\(m = 1\) indicates the layer is preserved, while \(m = 0\) means it is entirely dropped. Then \Cref{eq:pruning normal loss} can be transformed into finding an optimal binary mask vector $\mathcal{M}$ through \Cref{eq:loss}.
\begin{equation}
  \min \;\mathbb{E}_{(x,y)\sim\mathcal{D}}\bigl[\mathcal{L}\big(\mathcal{H}(x;\mathcal{M}\big),y\bigr],
  \label{eq:loss}
\end{equation}
\begin{figure}[t]
        \centering
        \includegraphics[width=0.99\linewidth]{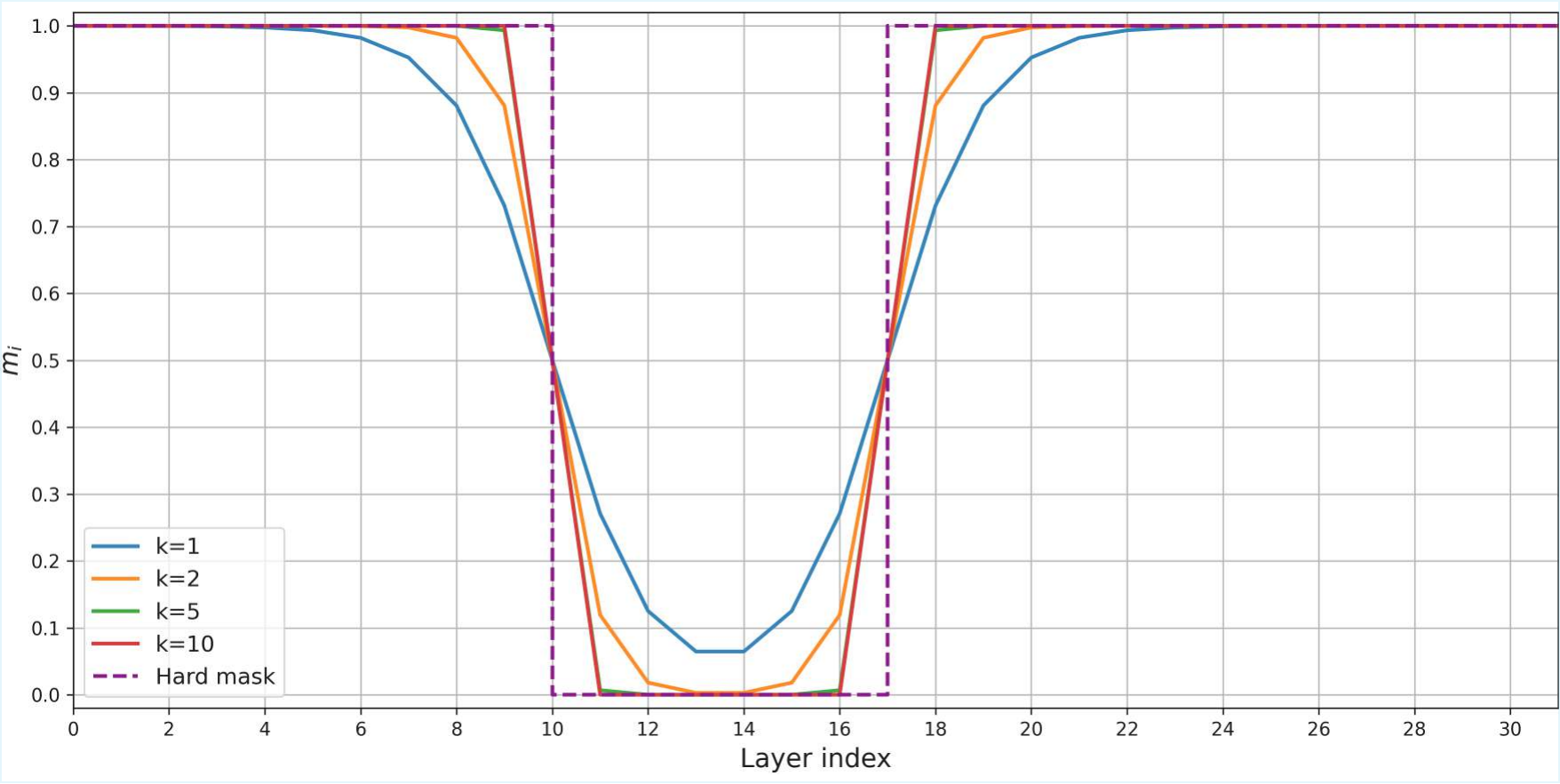}
    \caption{Visualization of the differentiable concave gating algorithm.}
    \label{fig:curves}
\end{figure}
However, directly applying the mask on \Cref{eq:standard-block} by
\begin{equation}
  o_i = m_i \cdot \mathrm{TransformerLayer}_i(o_{i-1}),
  \label{eq:naive-mask}
\end{equation}
would zero out the representations for all pruned layers, destroying gradients and stopping information flow. To avoid this, we introduce a residual connection in \Cref{eq:standard-block}:
\begin{equation}
  o_i = m_i \cdot \mathrm{TransformerLayer}_i(o_{i-1}) + (1-m_i) \cdot  o_{i-1}.
  \label{eq:skip-mask}
\end{equation}
By doing this, the pruned layers will be skipped directly without affecting the information flow and gradient propagation. In order to make the binary mask vector can be optimized by gradient, we further introduce a differentiable concave gating algorithm.
\begin{equation}
m_i=1-\frac{e^{k(i-a)}}{1+e^{k(i-a)}} \cdot \frac{1}{1+e^{k(i-a-n+1)}},
\label{eq:differentiable concave gating}
\end{equation}
where $k$ is a steepness hyperparameter and $i$ is the layer index. Unless otherwise specified, we set $k=5$ by default and present an ablation study on $k$ in \Cref{sec:ablation study}. Under this construction, when $k$ is large enough, \Cref{eq:differentiable concave gating} closely approximates a hard mask (see \Cref{fig:curves}):
\begin{itemize}
  \item For \(i<a\) or \(i>a+n-1\), one sigmoid saturates to 0, the other to 1, so \(m_i\approx1\) (layer kept).
  \item For \(a\le i\le a+n-1\), both sigmoids approach 1, so \(m_i\approx0\) (layer pruned).
\end{itemize}
Since $m_i$ consists of two sigmoid functions, we can compute the closed-form gradient and optimize it by backpropagation. 

\subsection{Mask Optimization via Difference Minimization}
Building upon the differentiable mask introduced above, the next step is to optimize the mask to identify the optimal layers for pruning. To this end, we formulate a learning objective aimed at minimizing the distributional discrepancy between the original model $\mathcal{H}$ and the soft-pruned model $\mathcal{H}_\mathrm{prune}$, drawing inspiration from knowledge distillation~\cite{yang2024learning,bao2023post,bao2024teacher}. Let \(z(x)\in\mathbb{R}^C\) and \(z_{\mathrm{prune}}(x)\in\mathbb{R}^C\) denote the logits produced by \(\mathcal{H}\) and $\mathcal{H}_\mathrm{prune}$ respectively for input \(x\). Their corresponding probability distributions are defined as:
\begin{equation}
    \begin{aligned}
        P_{\mathrm{full}}(\,\cdot\mid x)&=\operatorname{softmax}\!\bigl(z(x)\bigr),\\
P_{\mathrm{prune}}(\,\cdot\mid x)&=\operatorname{softmax}\!\bigl(z_{\mathrm{prune}}(x)\bigr).
    \end{aligned}
\end{equation}

The mask are then optimized by minimizing the KL divergence:
\begin{equation}
\mathbb{E}_{x\sim\mathcal{D}_{ca}}\bigl[\mathrm{KL}\bigl(P_{\mathrm{full}}(\cdot\mid x)\,\Vert\,P_{\mathrm{prune}}(\cdot\mid x)\bigr)\bigr],
\label{eq:combined-loss}
\end{equation}
where $\mathcal{D}_{ca}$ is the calibration set. The gradients $\partial \mathcal{L}/\partial a$ is computed via the chain rule through $m_i$ and backpropagated to update $a$ efficiently. After convergence in the continuous domain, we round $a$ to obtain an integer prune window and remove the corresponding layers.

\subsection{Fast Recovery with Cutoff Endpoint Tuning}
After obtaining the pruned model, fine-tuning is crucial to recover the performance drop caused by layer removal. Unlike traditional LLM pruning methods~\citep{kim2024shortened,men2024shortgpt}, which typically adopt parameter-efficient fine-tuning techniques such as LoRA~\cite{hu2022lora}, we propose a novel and efficient method called \textbf{Cutoff Endpoint Tuning}. The core idea of our method stems from the observation that removing a consecutive block of layers (e.g., from layer $a$ to layer $a+n-1$) creates a "cutoff" in the model, disrupting the information flow between the previous layer $a-1$ and the next layer $a+n$ in the pruned model. To effectively restore the model's functional integrity, we assume that the most critical adjustments occur at the direct interface of the removed blocks. Therefore, rather than fine-tuning all remaining parameters of the pruned model, our approach only fine-tunes the full parameters of the two layers adjacent to the pruned segment (i.e., the cutoff endpoint).

Formally, let the pruned model $\mathcal{H}_\mathrm{prune}$ be derived from the original model $\mathcal{H}$ by removing a contiguous set of layers from index $a$ to $a+n-1$. During cutoff endpoint tuning, only the weight parameters $\Theta_{\mathrm{tune}} = \{ W_{a-1}, W_{a+n} \}$ of layer $a-1$ and $a+n$ are updated, while all other parameters $\Theta_{\mathrm{fixed}}$ in $\mathcal{H}_\mathrm{prune}$ are kept frozen. Then the fine-tuning process is formulated as the following constrained optimization problem:
\begin{equation}
\min_{\Theta_{\mathrm{tune}}} \mathbb{E}_{(x,y) \sim \mathcal{D}_{ft}} \left[ \mathcal{L}\left( \mathcal{H}_{\mathrm{prune}}(x; \Theta_{\mathrm{tune}}, \Theta_{\mathrm{fixed}}), y \right) \right],
\end{equation}
where $\mathcal{D}_{ft}$ is the fine-tuning dataset. Finally, we obtain the fine-tuned pruned model.

\begin{table*}[t]
  \centering
  \caption{Zero-shot downstream task performance of various models using different pruning methods. The best results are marked in \textbf{boldface}.}
  \resizebox{1\textwidth}{!}{
    \begin{tabular}{c|c|c|ccccccc|c}
    \toprule
    Model & \multicolumn{1}{c|}{Method} & \multicolumn{1}{c|}{Pruning Rate} & \multicolumn{1}{c}{PIQA} & \multicolumn{1}{c}{HellaSwag} & \multicolumn{1}{c}{OpenbookQA} & \multicolumn{1}{c}{ARC-e} & \multicolumn{1}{c}{ARC-c} & \multicolumn{1}{c}{MMLU} & \multicolumn{1}{c|}{WinoGrande} & \multicolumn{1}{c}{Avg} \\
    \midrule
    \multirow{21}[6]{*}{LLaMA2-7B} & LLM-Pruner & 20.00\% & \textbf{96.08\%} & \textbf{89.34\%} & 89.59\% & 86.33\% & \textbf{86.34\%} & 57.33\% & 91.35\% & 86.71\% \\
          & FLAP  & 20.00\% & 94.29\% & 85.27\% & 89.59\% & 82.27\% & 78.77\% & 69.80\% & 90.78\% & 85.40\% \\
          & SliceGPT & 20.00\% & 77.50\% & 58.32\% & 69.23\% & 61.87\% & 61.43\% & 57.55\% & 86.24\% & 68.22\% \\
          & Bolaco & 20.00\% & 94.99\% & 84.73\% & \textbf{94.12\%} & \textbf{91.54\%} & 81.04\% & 75.05\% & 94.57\% & 88.88\% \\
          & SVD-LLM & 20.00\% & 82.39\% & 68.14\% & 85.97\% & 63.39\% & 60.15\% & 58.64\% & 89.98\% & 73.36\% \\
          & SoLA  & 20.00\% & 94.43\% & 84.19\% & 86.43\% & 88.08\% & 80.80\% & 74.62\% & 95.79\% & 87.43\% \\
          & CLP  & 20.00\% & \cellcolor[rgb]{ .867,  .922,  .969}92.76\% & \cellcolor[rgb]{ .867,  .922,  .969}87.35\% & \cellcolor[rgb]{ .867,  .922,  .969}86.62\% & \cellcolor[rgb]{ .867,  .922,  .969}89.57\% & \cellcolor[rgb]{ .867,  .922,  .969}86.23\% & \cellcolor[rgb]{ .867,  .922,  .969}\textbf{88.44\%} & \cellcolor[rgb]{ .867,  .922,  .969}\textbf{97.15\%} & \cellcolor[rgb]{ .867,  .922,  .969}\textbf{91.03\%} \\
\cmidrule{2-11}          & SliceGPT & 25.00\% & 84.69\% & 82.86\% & 81.00\% & 75.14\% & 75.00\% & 63.06\% & 92.09\% & 79.12\% \\
          & ShortGPT & 25.00\% & 75.41\% & 66.17\% & 79.19\% & 56.35\% & 67.39\% & 94.27\% & 89.02\% & 75.40\% \\
          & PruneNet & 25.00\% & 91.13\% & 82.02\% & /     & 80.64\% & 78.23\% & /     & 90.40\% & 84.48\% \\
          & DeeperLayers & 25.00\% & 85.19\% & 82.26\% & 85.07\% & 68.24\% & 76.09\% & 94.27\% & 93.70\% & 83.16\% \\
          & LLM-Pruner & 25.00\% & \textbf{94.98\%} & \textbf{90.38\%} & 87.33\% & 83.78\% & 80.87\% & 53.72\% & 91.07\% & 84.81\% \\
          & FINERCUT & 25.00\% & 94.60\% & 89.47\% & \textbf{90.05\%} & 83.92\% & \textbf{83.48\%} & 60.51\% & 94.29\% & 86.45\% \\
          & CLP  & 25.00\% & \cellcolor[rgb]{ .867,  .922,  .969}91.29\% & \cellcolor[rgb]{ .867,  .922,  .969}84.49\% & \cellcolor[rgb]{ .867,  .922,  .969}72.61\% & \cellcolor[rgb]{ .867,  .922,  .969}\textbf{84.89\%} & \cellcolor[rgb]{ .867,  .922,  .969}82.52\% & \cellcolor[rgb]{ .867,  .922,  .969}\textbf{94.42\%} & \cellcolor[rgb]{ .867,  .922,  .969}\textbf{99.09\%} & \cellcolor[rgb]{ .867,  .922,  .969}\textbf{87.05\%} \\
\cmidrule{2-11}          & LLM-Pruner & 30.00\% & \textbf{91.40\%} & 75.03\% & 83.71\% & 68.59\% & 68.45\% & 53.83\% & 78.61\% & 76.71\% \\
          & FLAP  & 30.00\% & 88.92\% & 74.53\% & 86.43\% & 74.08\% & 69.73\% & 58.42\% & 86.57\% & 78.74\% \\
          & SliceGPT & 30.00\% & 70.27\% & 46.33\% & 63.35\% & 52.44\% & 53.12\% & 56.67\% & 78.50\% & 60.46\% \\
          & Bolaco & 30.00\% & 90.88\% & 70.61\% & \textbf{87.78\%} & 78.82\% & 66.53\% & 61.27\% & 85.28\% & 82.68\% \\
          & SVD-LLM & 30.00\% & 75.91\% & 55.12\% & 76.92\% & 58.14\% & 54.98\% & 55.80\% & 83.96\% & 68.42\% \\
          & SoLA  & 30.00\% & 88.51\% & 75.22\% & 82.35\% & 79.38\% & 70.66\% & 60.61\% & 90.56\% & 82.99\% \\
          & CLP  & 30.00\% & \cellcolor[rgb]{ .867,  .922,  .969}89.34\% & \cellcolor[rgb]{ .867,  .922,  .969}\textbf{80.24\%} & \cellcolor[rgb]{ .867,  .922,  .969}82.17\% & \cellcolor[rgb]{ .867,  .922,  .969}\textbf{81.48\%} & \cellcolor[rgb]{ .867,  .922,  .969}\textbf{81.53\%} & \cellcolor[rgb]{ .867,  .922,  .969}\textbf{73.65\%} & \cellcolor[rgb]{ .867,  .922,  .969}\textbf{94.87\%} & \cellcolor[rgb]{ .867,  .922,  .969}\textbf{83.33\%} \\
\midrule    
        \multirow{21}[6]{*}{LLaMA2-13B} & LLM-Pruner & 20.00\% & \textbf{96.97\%} & 89.74\% & 96.49\% & 86.69\% & 90.09\% & 41.16\% & 83.79\% & 83.47\% \\
          & FLAP  & 20.00\% & 93.98\% & 87.13\% & 89.47\% & 85.17\% & 79.51\% & 74.37\% & 92.72\% & 86.12\% \\
          & SliceGPT & 20.00\% & 77.40\% & 59.56\% & 84.65\% & 60.20\% & 64.92\% & 55.96\% & 87.61\% & 66.43\% \\
          & Bolaco & 20.00\% & 95.55\% & 88.10\% & \textbf{98.25\%} & 91.65\% & 86.92\% & 78.34\% & 90.86\% & 90.85\% \\
          & SVD-LLM & 20.00\% & 89.04\% & 75.44\% & 88.60\% & 80.27\% & 74.65\% & 62.45\% & 94.35\% & 82.50\% \\
          & SoLA  & 20.00\% & 93.98\% & 84.81\% & 94.74\% & 89.35\% & 82.81\% & 83.21\% & 96.19\% & 90.91\% \\
          & CLP  & 20.00\% & \cellcolor[rgb]{ .867,  .922,  .969}96.55\% & \cellcolor[rgb]{ .867,  .922,  .969}\textbf{91.35\%} & \cellcolor[rgb]{ .867,  .922,  .969}95.14\% & \cellcolor[rgb]{ .867,  .922,  .969}\textbf{96.01\%} & \cellcolor[rgb]{ .867,  .922,  .969}\textbf{91.15\%} & \cellcolor[rgb]{ .867,  .922,  .969}\textbf{97.45\%} & \cellcolor[rgb]{ .867,  .922,  .969}\textbf{98.58\%} & \cellcolor[rgb]{ .867,  .922,  .969}\textbf{95.46\%} \\
\cmidrule{2-11}          & SliceGPT & 25.00\% & 73.68\% & 71.26\% & 78.76\% & 57.12\% & 59.59\% & 49.64\% & 86.03\% & 76.22\% \\
          & ShortGPT & 25.00\% & 76.75\% & 68.68\% & 78.32\% & 58.03\% & 65.92\% & 77.45\% & 87.71\% & 78.00\% \\
          & PruneNet & 25.00\% & 95.60\% & \textbf{88.73\%} & /     & 86.42\% & 81.98\% & /     & 89.73\% & 89.05\% \\
          & DeeperLayers & 25.00\% & 79.34\% & 74.43\% & 73.89\% & 59.33\% & 68.37\% & 83.82\% & 91.76\% & 75.92\% \\
          & LLM-Pruner & 25.00\% & 92.62\% & 86.49\% & \textbf{92.04\%} & 72.93\% & 72.24\% & 43.27\% & 80.73\% & 78.06\% \\
          & FINERCUT & 25.00\% & 84.38\% & 82.33\% & 78.76\% & 66.97\% & 66.53\% & 83.64\% & 93.16\% & 79.86\% \\
          & CLP  & 25.00\% & \cellcolor[rgb]{ .867,  .922,  .969}\textbf{95.93\%} & \cellcolor[rgb]{ .867,  .922,  .969}88.07\% & \cellcolor[rgb]{ .867,  .922,  .969}82.70\% & \cellcolor[rgb]{ .867,  .922,  .969}\textbf{93.06\%} & \cellcolor[rgb]{ .867,  .922,  .969}\textbf{90.25\%} & \cellcolor[rgb]{ .867,  .922,  .969}\textbf{97.16\%} & \cellcolor[rgb]{ .867,  .922,  .969}\textbf{96.63\%} & \cellcolor[rgb]{ .867,  .922,  .969}\textbf{91.97\%} \\
\cmidrule{2-11}          & LLM-Pruner & 30.00\% & 91.01\% & 76.68\% & \textbf{90.79\%} & 70.69\% & 69.26\% & 41.34\% & 79.87\% & 75.34\% \\
          & FLAP  & 30.00\% & 90.06\% & \textbf{78.63\%} & 85.96\% & 79.40\% & 75.87\% & 59.93\% & 88.14\% & 80.36\% \\
          & SliceGPT & 30.00\% & 70.58\% & 48.19\% & 69.30\% & 52.81\% & 53.29\% & 48.92\% & 79.55\% & 58.53\% \\
          & Bolaco & 30.00\% & 90.11\% & 72.70\% & 87.28\% & 84.77\% & 79.74\% & 61.91\% & 88.87\% & 83.01\% \\
          & SVD-LLM & 30.00\% & 81.53\% & 60.45\% & 82.46\% & 65.37\% & 61.10\% & 51.62\% & 88.14\% & 71.85\% \\
          & SoLA  & 30.00\% & 90.32\% & 77.30\% & 88.16\% & 84.73\% & 74.81\% & 71.12\% & 92.93\% & 85.20\% \\
          & CLP  & 30.00\% & \cellcolor[rgb]{ .867,  .922,  .969}\textbf{93.12\%} & \cellcolor[rgb]{ .867,  .922,  .969}68.37\% & \cellcolor[rgb]{ .867,  .922,  .969}65.16\% & \cellcolor[rgb]{ .867,  .922,  .969}\textbf{94.36\%} & \cellcolor[rgb]{ .867,  .922,  .969}\textbf{88.17\%} & \cellcolor[rgb]{ .867,  .922,  .969}\textbf{112.39\%} & \cellcolor[rgb]{ .867,  .922,  .969}\textbf{102.28\%} & \cellcolor[rgb]{ .867,  .922,  .969}\textbf{89.13\%} \\
    \bottomrule
    \end{tabular}}
  \label{tabs:sota}%
\end{table*}%

\section{Experiments}
\label{sec:Experiments}
In this section, we first compare our proposed method, CLP, with several state-of-the-art pruning methods under various pruning ratios to demonstrate its effectiveness. 
% Next, we evaluate the performance of CLP under various pruning ratios. 
Subsequently, we further conduct pruning experiments on more advanced (LLaMA3.1-8B-Instruct) and larger (LLaMA3-70B) models to demonstrate the generalization of our method. Finally, we conduct comprehensive ablation experiments to evaluate the effectiveness of each component in CLP.

\subsection{Experimental Settings}
\label{sec:settings}
\textbf{Models.} To evaluate the universality of our method, we conduct extensive experiments on $6$ popular open-source LLMs. These models span two major architecture (LLaMA and Qwen) and range in size from 7B to 70B parameters, including LLaMA2-7B~\citep{touvron2023llama}, LLaMA2-13B~\citep{touvron2023llama}, LLaMA3.1-8B-Instruct~\citep{grattafiori2024llama}, LLaMA3-70B~\citep{grattafiori2024llama}, Qwen1.5-7B~\cite{qwen} and Qwen1.5-14B~\cite{qwen}.

\textbf{Datasets.} To evaluate the effectiveness of CLP, we follow the evaluation of ~\cite{ma2023llm} to perform zero-shot task classification on $7$ common sense reasoning datasets, including MMLU~\citep{hendryckstest2021}
% , CMMLU~\citep{li2023cmmlu}
, PIQA~\citep{bisk2020piqa}, HellaSwag~\citep{zellers2019hellaswag}, WinoGrande~\citep{sakaguchi2021winogrande}, ARC-easy~\citep{clark2018think}, ARC-challenge~\citep{clark2018think} and OpenbookQA~\citep{mihaylov2018can}. All evaluations are conducted with lm-evaluation-harness~\citep{eval-harness}. 

\textbf{Baselines.} We compare CLP with $13$ state-of-the-art LLM pruning methods, including LLM-Pruner~\citep{ma2023llm}, ShortGPT~\citep{men2024shortgpt}, FLAP~\citep{an2024fluctuation}, SliceGPT~\citep{ashkboos2024slicegpt}, Shortened LLaMA~\citep{kim2024shortened}, SLEB~\citep{song2024sleb}, Bolaco~\citep{ji2024adaptive}, SVD-LLM~\citep{wang2024svd}, Reverse-order~\citep{lu2024reassessing}, PruneNet~\cite{sengupta2025you}, DeeperLayers~\cite{gromov2024unreasonable}, FINERCUT~\cite{zhang2024finercut} and SoLA~\citep{huang2025sola}. 

\textbf{Implementation Details.} In the pruning stage, we randomly select $3,000$ samples from C4 dataset~\cite{JMLR:v21:20-074} as the calibration set. Each sample is truncated to a sequence length of $256$. For the optimization of the differentiable mask, we set the learning rate to $0.5$, $a$ to $31$ and train for $1$ epoch, with $k=5$. During the fine-tuning stage, we utilize the cleaned version of Alpaca~\cite{taori2023stanford}, which comprises approximately $50k$ samples. We fine-tune the pruned model using cutoff endpoint tuning over $2$ epochs employing the AdamW optimizer with an initial learning rate of $1e-5$. Fine-tuning is conducted with a maximum sequence length of $256$ and a batch size of $64$. All experiments are conducted on $4$ A100 GPUs.

\begin{table*}[t]
  \centering
  \caption{Zero-shot downstream task performance of various model architectures using different pruning methods. The best results are marked in \textbf{boldface}, and the sub-optimal ones are \underline{underlined}.}
  \resizebox{1\textwidth}{!}{
    \begin{tabular}{c|c|c|ccccccc|c}
    \toprule
    Model & Method & Pruning Rate & PIQA  & HellaSwag & OpenbookQA & ARC-e & ARC-c & MMLU  & WinoGrande & Avg \\
    \midrule
    \multirow{5}[2]{*}{LLaMA3.1-8B-It} & Shortened LLaMA & 25.00\% & \textbf{95.51\%} & 83.43\% & 78.11\% & \textbf{89.01\%} & 73.58\% & 49.48\% & 78.58\% & 78.24\% \\
          & ShortGPT & 25.00\% & 89.85\% & 71.00\% & 59.76\% & 74.57\% & 54.94\% & 35.52\% & 73.13\% & 65.54\% \\
          & Reverse-order & 25.00\% & 87.67\% & 67.70\% & \textbf{86.98\%} & 75.34\% & \underline{77.06\%} & \underline{93.20\%} & 84.69\% & \underline{81.80\%} \\
          & SLEB  & 25.00\% & \underline{94.82\%} & \underline{84.15\%} & 79.29\% & 85.10\% & 74.74\% & 63.26\% & \underline{86.61\%} & 81.14\% \\
          & CLP  & 25.00\% & \cellcolor[rgb]{ .867,  .922,  .969}92.02\% & \cellcolor[rgb]{ .867,  .922,  .969}\textbf{86.41\%} & \cellcolor[rgb]{ .867,  .922,  .969}\underline{86.39\%} & \cellcolor[rgb]{ .867,  .922,  .969}\underline{87.88\%} & \cellcolor[rgb]{ .867,  .922,  .969}\textbf{87.45\%} & \cellcolor[rgb]{ .867,  .922,  .969}\textbf{97.27\%} & \cellcolor[rgb]{ .867,  .922,  .969}\textbf{89.61\%} & \cellcolor[rgb]{ .867,  .922,  .969}\textbf{89.58\%} \\
    \midrule
    \multirow{5}[2]{*}{Qwen1.5-7B} & LaCo  & 20.97\% & 88.87\% & 73.26\% & /     & 75.43\% & 77.00\% & /     & 88.23\% & 80.56\% \\
          & ShortGPT & 20.97\% & 85.03\% & 54.60\% & /     & \textbf{87.15\%} & 66.99\% & /     & 75.05\% & 73.76\% \\
          & RM    & 20.97\% & 87.77\% & 76.53\% & /     & 70.14\% & 75.41\% & /     & \underline{93.47\%} & 80.66\% \\
          & BlockPruner & 21.83\% & \underline{90.52\%} & \underline{77.11\%} & /     & \underline{86.39\%} & \underline{78.01\%} & /     & 83.60\% & \underline{83.13\%} \\
          & CLP  & 20.97\% & \cellcolor[rgb]{ .867,  .922,  .969}\textbf{91.88\%} & \cellcolor[rgb]{ .867,  .922,  .969}\textbf{83.86\%} & \cellcolor[rgb]{ .867,  .922,  .969}/ & \cellcolor[rgb]{ .867,  .922,  .969}80.45\% & \cellcolor[rgb]{ .867,  .922,  .969}\textbf{87.15\%} & \cellcolor[rgb]{ .867,  .922,  .969} /& \cellcolor[rgb]{ .867,  .922,  .969}\textbf{94.99\%} & \cellcolor[rgb]{ .867,  .922,  .969}\textbf{87.67\%} \\
    \midrule
    \multirow{5}[2]{*}{Qwen1.5-14B} & LaCo  & 22.25\% & 89.58\% & 75.76\% & /     & 78.42\% & 72.41\% & /     & 82.67\% & 79.77\% \\
          & ShortGPT & 22.25\% & 83.99\% & 52.99\% & /     & 74.07\% & 61.71\% & /     & 75.51\% & 69.65\% \\
          & RM    & 22.25\% & 73.37\% & 45.54\% & /     & 55.62\% & 74.05\% & /     & 79.31\% & 65.58\% \\
          & BlockPruner & 23.72\% & \textbf{94.20\%} & \textbf{84.27\%} & /     & \textbf{86.90\%} & \textbf{83.13\%} & /     & \underline{87.13\%} & \textbf{87.13\%} \\
          & CLP  & 22.25\% & \cellcolor[rgb]{ .867,  .922,  .969}\underline{90.34\%} & \cellcolor[rgb]{ .867,  .922,  .969}\underline{80.80\%} & \cellcolor[rgb]{ .867,  .922,  .969}/ & \cellcolor[rgb]{ .867,  .922,  .969}\underline{82.07\%} & \cellcolor[rgb]{ .867,  .922,  .969}\underline{81.36\%} & \cellcolor[rgb]{ .867,  .922,  .969}/ & \cellcolor[rgb]{ .867,  .922,  .969}\textbf{91.21\%} & \cellcolor[rgb]{ .867,  .922,  .969}\underline{85.16\%} \\
    \bottomrule
    \end{tabular}}
  \label{tab:more arch}%
\end{table*}%

% Table generated by Excel2LaTeX from sheet 'Sheet4'
\begin{table*}[htbp]
  \centering
  \caption{Zero-shot downstream task performance of LLaMA3-70B using different pruning methods. The best results are marked in \textbf{boldface}, and the sub-optimal ones are \underline{underlined}.}
    \resizebox{1\textwidth}{!}{
    \begin{tabular}{c|c|c|cccccccc|c}
    \toprule
    Model & Method & Pruning Rate & PIQA  & HellaSwag & OpenbookQA & ARC-e & ARC-c & MMLU  & CMMLU & WinoGrande & Avg \\
    \midrule
    \multirow{7}[2]{*}{LLaMA3-70B} & SliceGPT & 20.00\% & \underline{95.51\%} & 83.43\% & 78.11\% & 89.01\% & 73.58\% & 49.48\% & 49.14\% & 78.58\% & 75.81\% \\
          & ShortGPT & 20.00\% & 89.85\% & 71.00\% & 59.76\% & 74.57\% & 54.94\% & 35.52\% & 44.99\% & 73.13\% & 64.82\% \\
          & PIP   & 20.00\% & 87.67\% & 67.70\% & \underline{86.98\%} & 75.34\% & 77.06\% & 93.20\% & 98.30\% & 84.69\% & 83.66\% \\
          & SLEB  & 20.00\% & 94.82\% & 84.15\% & 79.29\% & 85.10\% & 74.74\% & 63.26\% & 60.22\% & 86.61\% & 79.61\% \\
          & Reverse-order & 20.00\% & 92.02\% & 86.41\% & 86.39\% & 87.88\% & \underline{87.45\%} & 97.27\% & \textbf{99.73\%} & 89.61\% & 91.05\% \\
          & CLP  & 20.00\% & \cellcolor[rgb]{ .867,  .922,  .969}\textbf{97.42\%} & \cellcolor[rgb]{ .867,  .922,  .969}\textbf{94.36\%} & \cellcolor[rgb]{ .867,  .922,  .969}\textbf{88.36\%} & \cellcolor[rgb]{ .867,  .922,  .969}\textbf{96.08\%} & \cellcolor[rgb]{ .867,  .922,  .969}\textbf{94.64\%} & \cellcolor[rgb]{ .867,  .922,  .969}\textbf{99.31\%} & \cellcolor[rgb]{ .867,  .922,  .969}96.64\% & \cellcolor[rgb]{ .867,  .922,  .969}\textbf{95.88\%} & \cellcolor[rgb]{ .867,  .922,  .969}\textbf{95.34\%} \\
          & CLP  & 30.00\% & \cellcolor[rgb]{ .867,  .922,  .969}94.12\% & \cellcolor[rgb]{ .867,  .922,  .969}\underline{87.75\%} & \cellcolor[rgb]{ .867,  .922,  .969}79.89\% & \cellcolor[rgb]{ .867,  .922,  .969}\underline{91.14\%} & \cellcolor[rgb]{ .867,  .922,  .969}86.29\% & \cellcolor[rgb]{ .867,  .922,  .969}\underline{98.32\%} & \cellcolor[rgb]{ .867,  .922,  .969}\underline{99.07\%} & \cellcolor[rgb]{ .867,  .922,  .969}\underline{93.29\%} & \cellcolor[rgb]{ .867,  .922,  .969}\underline{91.24\%} \\
    \bottomrule
    \end{tabular}}
  \label{tab:70b}%
\end{table*}%

\subsection{Comparison with the SoTA methods}
\label{sec:compare to sota}
To verify the effectiveness of our method, we conduct comparative experiments against existing SoTA baselines, including LLM-Pruner~\citep{ma2023llm}, ShortGPT~\citep{men2024shortgpt}, FLAP~\citep{an2024fluctuation}, SliceGPT~\citep{ashkboos2024slicegpt}, Bolaco~\citep{ji2024adaptive}, SVD-LLM~\citep{wang2024svd}, PruneNet~\cite{sengupta2025you}, DeeperLayers~\cite{gromov2024unreasonable}, FINERCUT~\cite{zhang2024finercut} and SoLA~\citep{huang2025sola}. However, existing pruning methods do not have a unified evaluation format (i.e., use different evaluation frameworks, lm-evaluation-harness~\citep{eval-harness} or OpenCompass~\cite{2023opencompass}), which makes fair comparison difficult. To address the issue of inconsistent evaluation methodologies, we employ the relative performance retention as a standardized metric for comparison. A higher retention rate indicates superior performance preservation. Specifically, we systematically evaluate the performance of the proposed method on LLaMA2-7B and LLaMA2-13B models with $3$ pruning rates of $20\%$, $25\%$, and $30\%$. As shown in \cref{tabs:sota}, experiments demonstrate that the proposed continuous layer pruning method consistently achieves excellent performance retention in zero-shot downstream tasks. For example, on the LLaMA2-7B model, when the pruning rate is $20\%$, our method achieves an average performance retention of approximately $91.0\%$, which is $2.15\%$ higher than the strongest baseline (Bolaco) and over $22.81\%$ higher than weaker baselines such as SliceGPT. On the larger LLaMA2-13B model, the advantage is even more pronounced: at a $20\%$ pruning rate, our method achieves an average of $95.46\%$, outperforming baselines by $4.55\%-29.03\%$. Under $25\%$ pruning rate, our method still performs well. For LLaMA2-7B, the average performance retention is $87.05\%$, which is $+0.6\%$ to $+11.65\%$ higher than existing baselines; for LLaMA2-13B, the average performance retention is about $92.0\%$, which is $+2.92\%$ to $+16.05\%$ higher than baselines. At a $30\%$ pruning rate, our method also maintains a lead: for LLaMA2-7B, the average performance retention is about $83.3\%$, with a lead of about $+0.3\%$ to $+22.9\%$; for LLaMA2-13B, the average performance retention is $89.13\%$, with a lead of $+3.9\%$ to $+30.6\%$ over the baselines. In summary, our method significantly outperforms all existing baselines in pruning performance retention under different pruning rates.

\textbf{Expanding to More Model Structures.} We further extend our validation to LLaMA3.1-8B-Instruct and the Qwen family (Qwen1.5-7B~\cite{qwen} and Qwen1.5-14B~\cite{qwen}) to demonstrate that our method can generalize beyond the LLaMA2 family. The experimental results are shown in \cref{tab:more arch}. In almost all benchmarks, our method significantly outperforms or is on par with other comparison methods in terms of maintaining model performance. Specifically, after pruning the LLaMA3.1-8B-It model by $25.00\%$, our method achieves an average performance retention rate of $89.58\%$, ranking first among all baselines. This result not only significantly outperforms baseline methods such as ShortGPT ($65.54\%$) and Shortened LLaMA ($78.24\%$), but also outperforms the next-best SLEB ($81.14\%$) and Reverse-order ($81.80\%$) methods by $8.44\%$ and $7.78\%$, respectively. For the Qwen1.5-7B model, our method once again achieves the top score with a relative performance retention of $87.67\%$ at a pruning rate of approximately $21\%$. This outperforms the next-best performer, BlockPruner ($83.13\%$), by over $4.5\%$ and significantly surpasses other methods, including ShortGPT ($73.76\%$), LaCo ($80.56\%$) and RM ($80.66\%$). For the larger Qwen1.5-14B model, our method achieves an average performance retention of $85.16\%$ at a pruning rate of approximately $22.25\%$, demonstrating strong competitiveness. Although BlockPruner leads with an average score of $87.13\%$ on this model, our method still significantly outperforms all other comparison methods, including LaCo ($79.77\%$), ShortGPT ($69.65\%$), and RM ($65.58\%$). Overall, the experiments strongly demonstrate the superiority of our proposed pruning method. It not only consistently achieves leading average performance across a variety of model structures and pruning rates, but also achieves top performance on most individual downstream tasks.

\textbf{Expanding to Larger Models.} To further validate the performance and scalability of our pruning method on very large models, we conduct experiments on the LLaMA3-70B model and compare it with several state-of-the-art pruning methods at different pruning rates. As shown in \cref{tab:70b}, the experimental results clearly demonstrate the superior performance of our method. Specifically, with all methods set to a $20.00\%$ pruning rate, our method achieves an average performance retention rate of $95.34\%$, significantly surpassing all compared baselines. Compared to the next-best performing Reverse-order method ($91.05\%$), our method achieves a significant advantage of over $4.29\%$. To explore the performance limit of our method, we increase the pruning rate to $30.00\%$. Even with a significantly higher compression rate than other methods, our method still achieves an average performance retention rate of $91.24\%$, outperforming the best performance of all other methods at a $20\%$ pruning rate. This result is very convincing, demonstrating the effectiveness of our pruning strategy, which can remove more model parameters ($10\%$ more) while retaining more performance than other baselines. Finally, our approach provides a very effective solution for efficiently compressing very large language models.

\begin{figure}[htbp]
        \centering
        \includegraphics[width=0.99\linewidth]{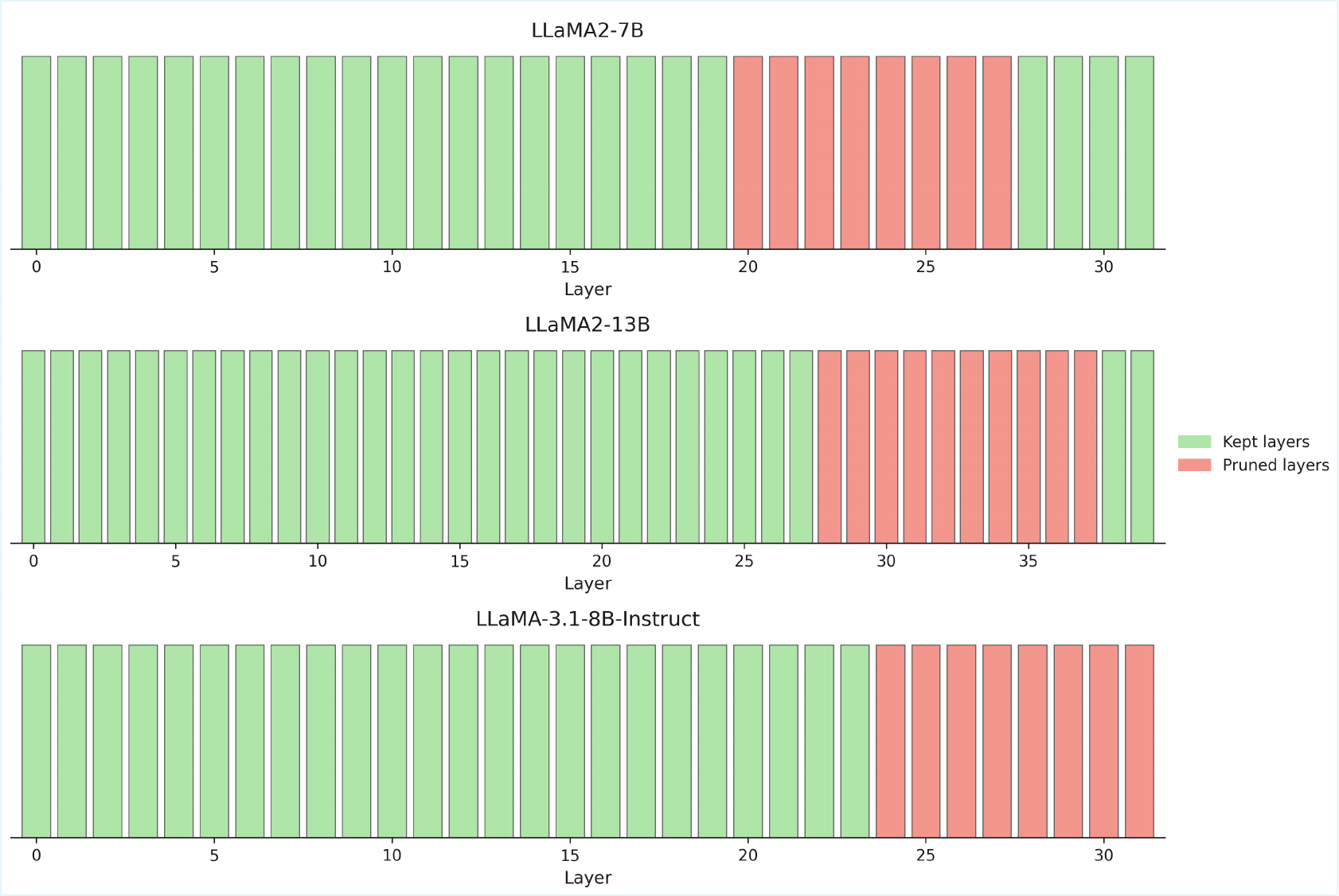}
    \caption{Visualization of pruned layers at $25\%$ pruning rate.}
    \label{fig:layer vis}
\end{figure}

\begin{figure}[htbp]
        \centering
        \includegraphics[width=0.99\linewidth]{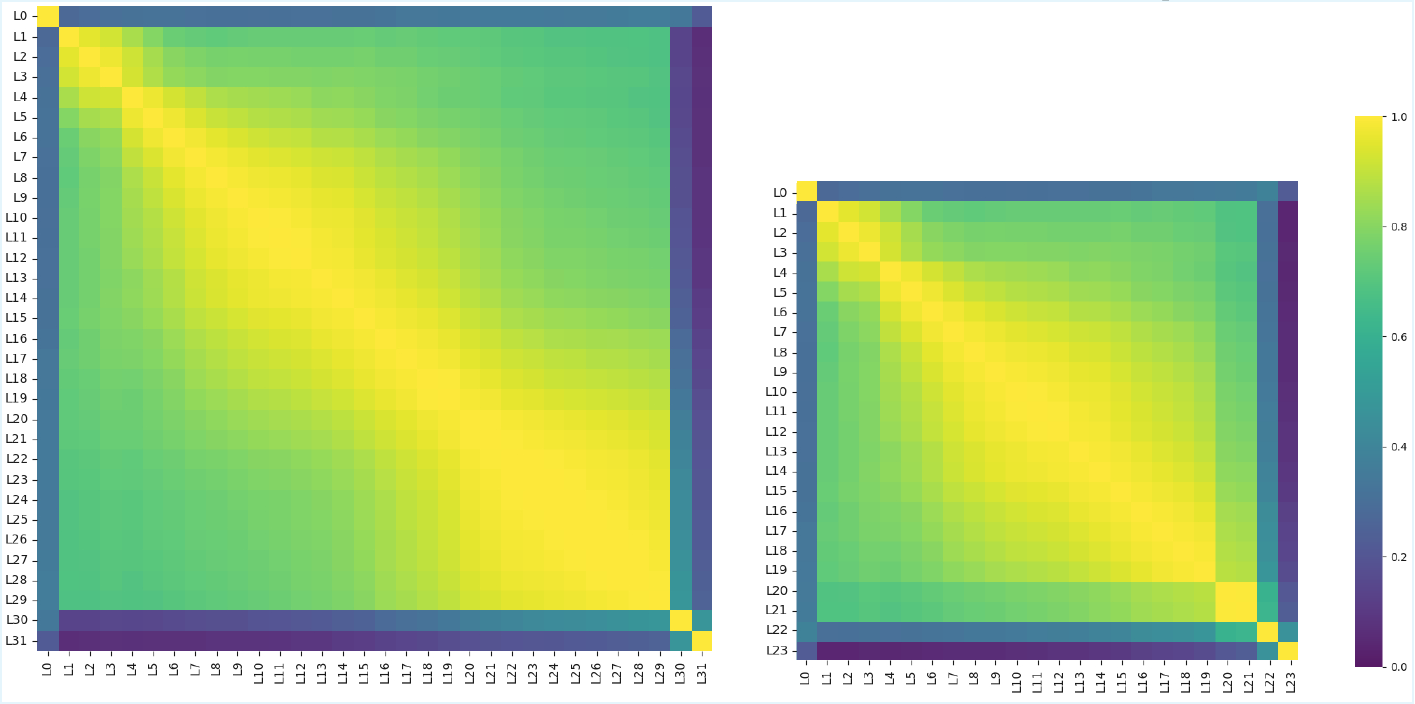}
    \caption{Layer similarity visualization. The original LLaMA2-7B model is on the left, and the pruned model obtained by cutting off $25\%$ of the layers using our method is on the right.}
    \label{fig:cka}
\end{figure}

\subsection{Additional Experiments.}
In this subsection, we conduct a series of additional experiments to analyze the effectiveness of the proposed method and the actual speedup effect.

\textbf{Visualization of Pruned Layers.} To further illustrate the effectiveness of our pruning method, we visualize the layers pruned by CLP on LLaMA2-7B, LLaMA2-13B, and LLaMA3.1-8B-Instruct at $25\%$ pruning rate. As shown in \cref{fig:layer vis}, our method predominantly removes layers in the medium-to-late stages of the model. This result aligns with and is empirically supported by existing research~\cite{skean2025layer} on the layer-wise functionality in LLMs.

\begin{table*}[t]
    \centering
    % \vspace{-4mm}
    \caption{Generated Examples from the original LLaMA-3.1-8B-It and our pruned LLaMA-3.1-8B-It.} \label{tab:Generated Examples_Llama-3.1-6.3B-It}
    \resizebox{\linewidth}{!}{
    \tiny
    \begin{tabular}{p{1.6cm}|p{9cm}}
        \toprule
        Model & Generated Sentences \\
        \midrule
        \multirow{5}{*}{\parbox{1.6cm}{LLaMA-3.1-8B-It}} & \underline{To be or not to be, that is the question}, or so we tell ourselves. But what if there's no question at stake? What if our choices are predetermined, the result of a complex interplay between genetics, environment, upbringing, culture, society... We like the idea that we're free agents, capable of making decisions that shape our own destiny, but what evidence do we truly have to support this notion? Is it merely an illusion, created by the comfort of believing we hold sway over our lives? \\
        \midrule 
        \multirow{5}{*}{\parbox{1.6cm}{LLaMA-3.1-8B-It-prune}} & \underline{To be or not to be, that is the question} asked by the very existence itself. Is it a matter of choice or simply a given? We often think of ourselves as having control over our lives, as being able to choose our path in life, but is it really true? Or are we merely floating along with the flow of events, taking turns and turns until we reach our destination? And what does it mean to exist anyway? Is existence a state, an state in which we are living breathing beings. \\
        \bottomrule
    \end{tabular}
    }
\end{table*}

\textbf{Inter-layer Similarity Analysis of Pruned Models.} To further explore the impact of our pruning method on the internal structure of LLMs, we conduct a visualization analysis experiment. Specifically, we select LLaMA2-7B as the base model and pruned $25\%$ of its layers using our method. We then calculate the inter-layer feature similarity of the original model and the pruned model using centered kernel alignment~\cite{kornblith2019similarity}, and visualize these similarity matrices as heatmaps, as shown in \cref{fig:cka}. In the original LLaMA2-7B model on the left, we can clearly see a bright yellow region in its deep layers, indicating significant information redundancy in these layers. After pruning using our method, the model on the right is not only smaller in size, but more importantly, this highly redundant deep region has been successfully removed. This demonstrates that our method does not simply truncate the model, but rather analyzes the functional relationships between layers to precisely locate and remove overlapping portions that have a low marginal contribution to the overall model performance. This is why our pruning method is able to significantly compress the model while retaining good performance to the greatest extent possible.

\begin{figure}[t]
        \centering
        \includegraphics[width=0.99\linewidth]{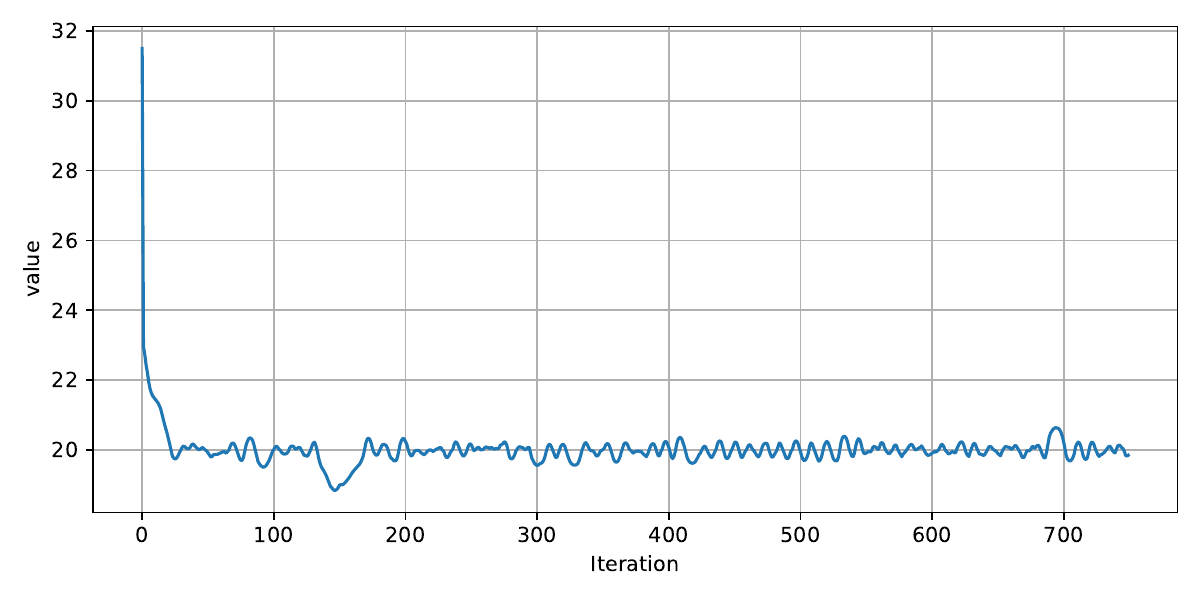}
    \caption{Visualization of the dynamic changes of $a$. The experiment is conducted on the LLaMA2-7B with the $25\%$ pruning rate.}
    \label{fig:vis a}
\end{figure}

\textbf{Analysis of Mask Parameter $a$'s Dynamics.} As shown in \cref{eq:differentiable concave gating}, $a$ is the learnable parameter that represents the starting index of the consecutive layer segments to be pruned. This experiment aims to visualize the optimization process of the differentiable mask parameter $a$. We conduct experiments on the LLaMA2-7B model under a $25\%$ pruning rate. Throughout training, the parameter $a$ is optimized via gradient descent to minimize the KL divergence loss between the original and soft-pruned models. As shown in \cref{fig:vis a}, the value of $a$ converges to a stable value after an initial exploration of the model depth. This convergence indicates that the gradient-based optimization effectively determines the locations for pruning.

\textbf{Case Study.} We provide some examples of sentences generated by the LLaMA-3.1-8B-It compressed using our pruning method in \cref{tab:Generated Examples_Llama-3.1-6.3B-It}. From the cases in \cref{tab:Generated Examples_Llama-3.1-6.3B-It}, the sentences generated by the pruned model are comparable to those generated by the original model, exhibiting fluency and relevance to the given topic. This also reflects from the side that the model obtained by our pruning method has strong reasoning ability.

\begin{figure*}[t]
        \centering
        \includegraphics[width=0.99\linewidth]{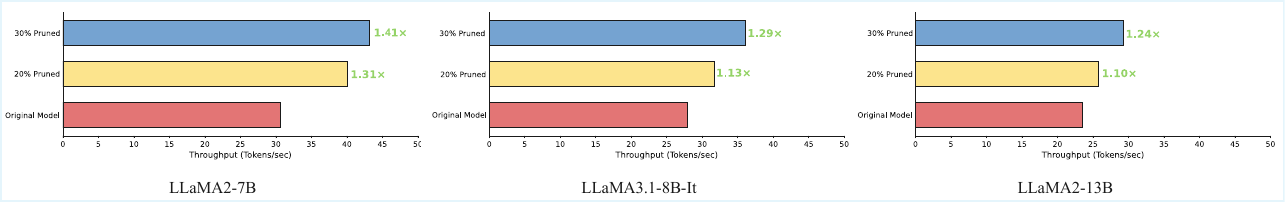}
    \caption{The actual speedup of pruned models obtained by our method.}
    \label{fig:inference}
\end{figure*}

% Table generated by Excel2LaTeX from sheet 'Sheet4'
\begin{table*}[htbp]
  \centering
  \caption{Further quantization using GPTQ on the pruned model. We conduct experiments on LLaMA2-7B and LLaMA2-13B at a $25\%$ pruning rate.}
  \resizebox{1\textwidth}{!}{
    \begin{tabular}{c|c|c|c|ccccccc|c}
    \toprule
    Model & Method & \#Param & Memory & \multicolumn{1}{c}{PIQA} & \multicolumn{1}{c}{HellaSwag} & \multicolumn{1}{c}{OpenbookQA} & \multicolumn{1}{c}{ARC-e} & \multicolumn{1}{c}{ARC-c} & \multicolumn{1}{c}{MMLU} & \multicolumn{1}{c|}{WinoGrande} & \multicolumn{1}{c}{Avg Acc} \\
    \midrule
    \multirow{3}[2]{*}{LLaMA2-7B (FP16)} & Dense & 6.74B & 12852.51 MB & 78.07 & 57.14 & 31.40 & 76.35 & 43.43 & 41.97 & 69.14 & 56.79  \\
          & CLP   & 5.12B & 9764.38 MB & 71.27 & 48.28 & 22.80 & 64.81 & 35.84 & 39.63 & 68.51 & 50.16  \\
          & CLP+GPTQ & 5.12B & 1270.02 MB & 70.02 & 47.01 & 23.00 & 62.25 & 34.56 & 34.04 & 66.77 & 48.24  \\
    \midrule
    \multirow{3}[2]{*}{LLaMA2-13B (FP16)} & Dense & 13.02B & 24945.81 MB & 78.61 & 60.68 & 37.00 & 78.87 & 48.12 & 52.50 & 72.77 & 61.22  \\
          & CLP   & 9.84B & 18865.61 MB & 75.41 & 53.44 & 30.60 & 73.40 & 43.43 & 51.01 & 70.32 & 56.80  \\
          & CLP+GPTQ & 9.84B & 2545.11 MB & 74.59 & 53.41 & 26.80 & 72.43 & 42.41 & 51.32 & 70.64 & 55.94  \\
    \bottomrule
    \end{tabular}}
  \label{tab:gptq}%
\end{table*}%

\textbf{Actual Speedup Effect of Pruning Models.} To quantify the practical speedup effect of our pruning method, we systematically test the inference speed of pruned models. We conduct experiments on $3$ models: LLaMA2-7B, LLaMA3.1-8B-It, and LLaMA2-13B. Specifically, for each model, we test the original model, the model pruned by $20\%$ using our method and the model pruned by $30\%$ using our method. Following previous studies, we evaluate the model's inference speed by measuring its throughput. This is measured in tokens generated per second (tokens/sec), with higher values indicating faster execution. As shown in \cref{fig:inference}, experimental results consistently show that our pruning method can bring significant inference speed improvements for all models, which are positively correlated with the pruning rate. For example, for the LLaMA2-7B model, $30\%$ pruning bring a speedup of up to $1.41\times$. On the larger LLaMA3.1-8B-It and LLaMA2-13B models, $30\%$ pruning also achieves inference speedups of $1.29\times$ and $1.24\times$, respectively, validating the effectiveness of our approach in achieving efficient and lightweight LLMs.

\textbf{Synergy with Post-Training Quantization.} To demonstrate the compatibility of our method with other mainstream compression techniques, we further apply GPTQ~\cite{frantar2022gptq} quantization to models pruned using our approach. We first prune $25\%$ of the layers from LLaMA2-7B and 13B using our method, subsequently apply 4-bit GPTQ quantization to the pruned models, and finally assess the performance of the final compressed models. As shown in \cref{tab:gptq}, while quantization introduces an expected additional performance degradation on LLaMA2-13B (a decrease of $0.86$) compared to the pruned but unquantized model, the quantized model achieves a significant reduction in memory usage ($2545.11$MB with GPTQ vs. $18865.61$MB without). Besides, LLaMA2-13B is reduced from $24945.81$MB to just $2545.11$MB, a nearly $86.51\%$ reduction, while still maintaining the $91.37\%$ average accuracy of the original dense model. The above experiments demonstrate that our approach can be used together with quantization to achieve strong two-stage compression for deploying large models under strict resource constraints.

\begin{table}[t]
  \centering
  \caption{Comparison of LoRA and cutoff endpoint tuning. Experiments are conducted on LLaMA2-7B and LLaMA2-13B at a $25\%$ pruning rate.}
  \resizebox{0.49\textwidth}{!}{
    \begin{tabular}{c|cccc}
    \toprule
    \multicolumn{1}{c|}{Model} & Method & Training Time & \multicolumn{1}{c}{Trainable Parameters} & \multicolumn{1}{c}{Avg Acc} \\
    \midrule
    \multirow{2}[2]{*}{LLaMA2-7B} & LoRA & \multicolumn{1}{c}{7977.0s} & 0.29\% & 43.21 \\
          & CLP & 5245.8s & 10.47\% & 47.51 \\
    \midrule
    \multirow{2}[2]{*}{LLaMA2-13B} & LoRA & \multicolumn{1}{c}{11821.8s} & 0.24\% & 51.74 \\
          & CLP & 9548.6s & 8.11\% & 54.14 \\
    \bottomrule
    \end{tabular}}
  \label{tab:LoRA vs our}%
\end{table}%

\begin{figure}[t]
        \centering
        \includegraphics[width=0.99\linewidth]{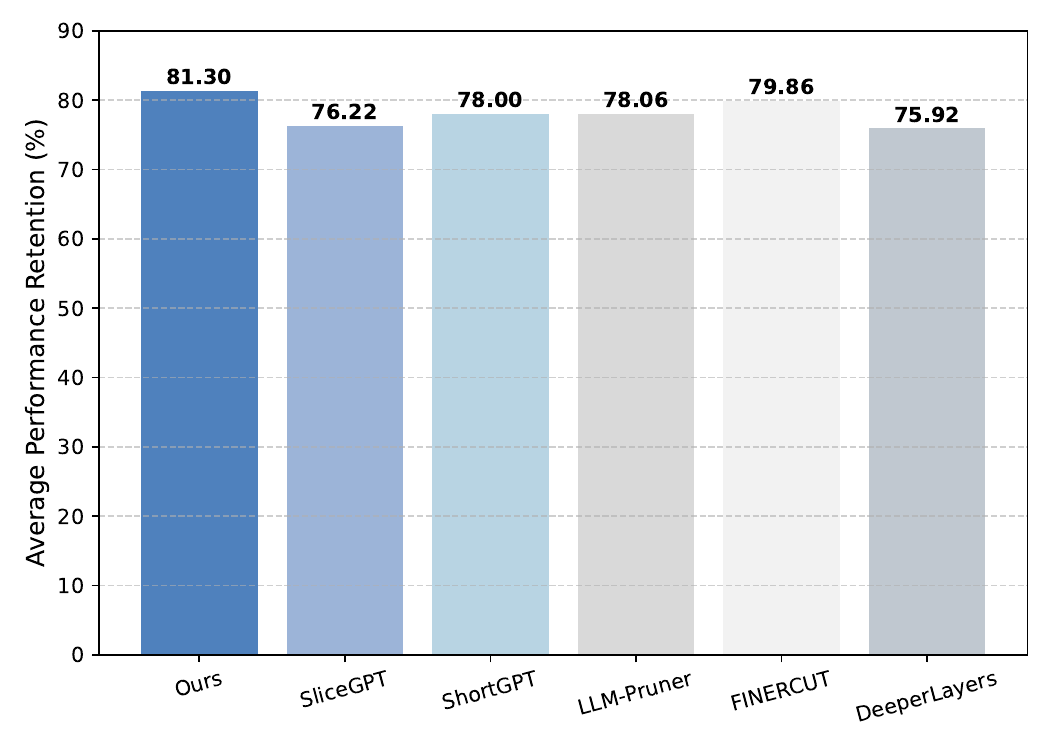}
    \caption{Verify the effectiveness of our pruning method without the cutoff endpoint tuning strategy. The experiment is conducted on the LLaMA2-13B with the $25\%$ pruning rate. All methods are fine-tuning with LoRA.}
    \label{fig:aba pruning}
\end{figure}

\subsection{Ablation Study}
\label{sec:ablation study}
In this subsection, we conduct systematic ablation experiments to demonstrate the effectiveness and reliability of our proposed method.

\textbf{LoRA vs. Cutoff Endpoint Tuning.} In this paper, we use cutoff endpoint tuning instead of LoRA for performance recovery of pruned models. Therefore, we conduct a comparative study between the two methods on LLaMA2-7B and LLaMA2-13B at a $25\%$ pruning rate. Specifically, we fine-tune the pruned model on the Alpaca-cleaned dataset using LoRA and cutoff endpoint tuning for $2$ epochs, and calculate the trainable parameters, training time, and the average performance on the $7$ datasets. As shown in \cref{tab:LoRA vs our}, while LoRA trains a very small proportion of parameters ($0.29\%$ for 7B, $0.24\%$ for 13B), our cutoff endpoint tuning achieves significantly higher average accuracy ($47.51$ vs. $43.21$ for 7B; $54.14$ vs. $51.74$ for 13B). Notably, our approach also significantly reduces training time. Compared to LoRA, cutoff endpoint tuning reduces training time by approximately $34\%$ on LLaMA2-7B and by approximately $19\%$ on LLaMA2-13B. These experimental results demonstrate that fine-tuning key connection layers adjacent to the pruned segments not only more effectively restores model capacity but is also more efficient than applying LoRA to the entire network. This highlights the advantages of targeted, architecture-design-based fine-tuning over general parameter-efficient fine-tuning methods in the context of structured pruning.

\textbf{Superiority of the Pruning Method Itself.} Our pruning method consists of two core components: (1) a differentiable mask for identifying layers to prune; and (2) a dedicated cutoff endpoint tuning strategy for performance recovery. In \cref{sec:compare to sota} and \cref{tab:LoRA vs our}, we have demonstrated the superiority of our pruning method and cutoff endpoint tuning strategy, respectively. Here, we want to verify that pruned models obtained solely with differentiable masks can achieve good performance without the specialized cutoff endpoint tuning strategy. To isolate and validate the contribution of the first component (i.e., the pruning algorithm itself), we conduct a controlled experiment in which all methods use the same LoRA fine-tuning to recover performance. The experiment is conducted on the LLaMA2-13B with the $25\%$ pruning rate. As shown in \cref{fig:aba pruning}, our method achieves an average performance retention of $81.30\%$, significantly outperforming all other baseline methods. Since the same fine-tuning strategy is used, the performance gap can be directly attributed to the advantage of our gradient-based mask search in identifying superior subnetworks. This ultimately demonstrates that the core of our pruning method is inherently more effective than existing baselines.

\textbf{Experimental Analysis of Hyperparameter $k$.} As mentioned in \cref{fig:curves}, $k$ is a hyperparameter that controls the steepness of our differentiable concave gating algorithm. Therefore, we conduct an ablation study to determine its impact on pruning performance. Specifically, we conduct experiments on LLaMA2-7B with the $25\%$ pruning rate and keep $a=31$ unchanged. We change the value of $k$ to $3$, $5$, or $10$ to simulate the change of steepness. As shown in \cref{tab:fixed_a_independent}, our method is not highly sensitive to the exact setting of $k$. 

\textbf{Experimental Analysis of Hyperparameter $a$.} In this paper, $a$ is a learnable parameter that represents the starting index of the consecutive layer segments to be pruned. As mentioned in \cref{sec:settings}, we set $a=31$ by default. Here, we analyze the sensitivity of the final performance of the pruned model to the initial value $a$. Specifically, we conduct experiments on LLaMA2-7B with the $25\%$ pruning rate. We keep $k=5$ unchanged and change $a$ to $10, 23$, and $31$ to simulate different initial positions for mask learning and fine-tuning. As shown in \cref{tab:fixed_k_independent}, regardless of the initial starting point, the optimization process always converges to a similar optimal pruning region. This shows that the gradient-based search is effective and not very sensitive to initialization.

\begin{minipage}[htbp]{0.23\textwidth} % 第一个表格的容器
    \begin{table}[H] % 使用[H]确保表格就出现在这里
        \centering
        \caption{Ablation study on $k$.}
        \label{tab:fixed_a_independent}
        \begin{tabular}{ccc}
        \toprule
        \textbf{a} & \textbf{k} & \textbf{Avg} \\
        \hline
        \multirow{3}{*}{31} & 3 & 86.89\% \\
                          & 5 & 87.05\% \\
                          & 10 & 87.32\% \\
        \bottomrule
        \end{tabular}
    \end{table}
\end{minipage}
\hfill % 水平弹性空间
\begin{minipage}[htbp]{0.23\textwidth} % 第二个表格的容器
    \begin{table}[H] % 使用[H]确保表格就出现在这里
        \centering
        \caption{Ablation study on $a$.}
        \label{tab:fixed_k_independent}
        \begin{tabular}{ccc}
        \toprule
        \textbf{k} & \textbf{a} & \textbf{Avg} \\
        \hline
        \multirow{3}{*}{5} & 10 & 86.79\% \\
                         & 23 & 87.14\% \\
                         & 31 & 87.05\% \\
        \bottomrule
        \end{tabular}
    \end{table}
\end{minipage}

\section{Conclusion}
\label{sec:Conclusion}
In this paper, we introduce CLP, a novel framework for continuous layer pruning of LLMs. Specifically, CLP utilizes a differentiable concave gating algorithm to automatically discover optimal pruning regions through gradient-based optimization. Subsequently, a targeted cutoff endpoint tuning strategy is employed to restore the performance of the pruned model by focusing on fine-tuning the key connection points created by pruning. Our extensive experiments demonstrate that CLP consistently outperforms existing pruning methods across multiple model families and scales, significantly reducing recovery time while maintaining superior performance. Furthermore, CLP is compatible with post-training quantization, achieving extremely high compression rates with only a small performance loss. This opens up new possibilities for deploying LLMs in resource-constrained environments.

% \clearpage
\bibliographystyle{IEEEtran}
\bibliography{reference}

\end{document}